%% file: Paper-1358.tex
\documentclass[runningheads]{llncs}
\usepackage[T1]{fontenc}
\usepackage{graphicx}
\usepackage{amsmath}
\usepackage{amssymb}
\usepackage{listings}
\usepackage[colorlinks,hyperindex,breaklinks]{hyperref}
\usepackage{multirow}
\usepackage{multicol}
\usepackage{caption}
\usepackage{pifont}
\usepackage{booktabs}
\usepackage{marvosym}

\newcommand{\ie}{\textit{i}.\textit{e}.}
\newcommand{\etal}{\textit{et al}.}
\newcommand{\eg}{\textit{e}.\textit{g}.}
\usepackage{lineno}
\newcommand{\cmark}{\ding{51}}
\newcommand{\xmark}{\ding{55}}

\begin{document}
% ----------------------------------------
\title{Aligning Medical Images with General Knowledge from Large Language Models}
% ----------------------------------------
\author{Xiao Fang$^{\dag 1}$,  %index{Fang, Xiao}
Yi Lin$^{\dag 1}$, %index{Lin, Yi}
Dong Zhang$^{2}$, %index{Zhang, Dong}  
Kwang-Ting Cheng$^{2}$, %index{Cheng, Kwang-Ting}
Hao Chen\textsuperscript{\Letter}$^{1,3, 4}$ %index{Chen, Hao}
}
\institute{$^{1}$Department of Computer Science and Engineering, HKUST, Hong Kong, China \\ 
$^{2}$Department of Electronic and Computer Engineering,  HKUST, Hong Kong, China \\
$^{3}$Department of Chemical and Biological Engineering, HKUST, Hong Kong, China \\ 
$^{4}$HKUST Shenzhen-Hong Kong Collaborative Innovation Research Institute, Futian, Shenzhen, China. \\
\email{jhc@cse.ust.hk}}
\titlerunning{Aligning Medical Images with Knowledge from Large Language Models}
\authorrunning{X. Fang and Y. Lin, \etal}

\maketitle          
\def\thefootnote{$\dag$}\footnotetext{Equal contribution; \Letter~corresponding author.}
% typeset the header of the contribution
%------------------------------------------
\input{Section/abstract}
%------------------------------------------
\input{Section/introduction}
%------------------------------------------
\input{Section/method}
%------------------------------------------
\input{Section/experiment}
%------------------------------------------
\input{Section/conclusion}
%------------------------------------------

\bibliographystyle{splncs04}
\bibliography{Paper-1358}
%------------------------------------------
\end{document}

%% file: Section/abstract.tex
\begin{abstract}
Pre-trained large vision-language models (VLMs) like CLIP have revolutionized visual representation learning using natural language as supervisions, and demonstrated promising generalization ability. In this work, we propose ViP, a novel visual symptom-guided prompt learning framework for medical image analysis, which facilitates general knowledge transfer from CLIP. ViP consists of two key components: a visual symptom generator (VSG) and a dual-prompt network. Specifically, VSG aims to extract explicable visual symptoms from pre-trained large language models, while the dual-prompt network utilizes these visual symptoms to guide the training on two learnable prompt modules, \ie, \emph{context prompt} and \emph{merge prompt}, which effectively adapts our framework to medical image analysis via large VLMs. 
Extensive experimental
results demonstrate that ViP can outperform state-of-the-art methods on two challenging datasets. The code is available at \url{https://github.com/xiaofang007/ViP}.
% ----------------------------------
\keywords{Prompt Learning \and Vision-Language Models \and Large Language Model \and Medical Image Analysis.}
\end{abstract}

%% file: Section/introduction.tex
\section{Introduction}
Medical image analysis plays a crucial role in healthcare, enabling non-invasive diagnosis and treatment of various medical conditions~\cite{lin2023boosting,lin2023rethinking,chen2022transmorph,you2021aligntransformer}. With the advent of deep learning techniques, computer-aided medical image analysis has achieved remarkable success in numerous scenarios. Current methods generally adopt the supervised learning paradigm which requires a large amount of labeled data for model training. However, this paradigm relies on manual annotation of medical images, which is time-consuming and labor-intensive~\cite{zheng2024exploring}.

The emergence of large vision language models (VLMs)~\cite{li2022blip,li2022grounded,li2023scaling} makes it possible to transfer knowledge from large-scale pre-trained models to task-specific medical image analysis models with limited data. One prominent example is Contrastive Language-Image Pre-training (CLIP)~\cite{radford2021learning}, which is pre-trained on $400$ million image-text pairs using contrastive learning. In detail, it comprises a vision and a text encoder that encodes an image and its corresponding text snippet into visual and textual embeddings, respectively. While CLIP has demonstrated great potential in transfer learning across diverse tasks in universal scenes, its direct applications to the medical domain raise challenges. This is because CLIP is pre-trained mainly on web-scraped data, which primarily comprises natural image-text pairs and lacks medical data due to privacy concerns, while the category texts of medical images tend to be abstract medical lexicons, which can be hard for CLIP to interpret. Inspired by recent work~\cite{menon2022visual,qin2022medical}, we propose to address the interpreting challenge by translating abstract medical lexicons to visual symptoms that are shared across natural and medical domains, such as color, shape and texture. In this way, VLMs can learn to align image features with visual features that are easily interpreted. This process also aligns with the diagnostic approach employed by medical professionals, who diagnose diseases based on related visual features observed in medical images.

In this paper, we propose ViP, a novel \textbf{Vi}sual symptom-guided \textbf{P}rompt learning framework that promotes general knowledge transfer of CLIP~\cite{radford2021learning}. The framework consists of two main components: A visual symptom generator (VSG) and a dual-prompt network. VSG queries pre-trained large language models (LLMs) to generate visual symptoms, which serve as text inputs for the dual-prompt network. The dual-prompt network enhances the generalization ability of CLIP by training two learnable prompt modules: \emph{context prompt} (CoP) and \emph{merge prompt} (MeP). CoP refines visual symptoms by incorporating medical task context while MeP aggregates text features of visual symptoms. The proposed framework is evaluated on two public datasets, including Pneumonia~\cite{kermany2018identifying} and Derm7pt~\cite{kawahara2018seven}. Extensive experimental results demonstrate that ViP outperforms state-of-the-art methods, highlighting the efficacy of each component in our framework.

The main contributions of our work are as follows: 1) We reveal the significant impact of LLMs on prompt engineering, showcasing their influence on enhancing interpretability and performance. 2) We propose ViP that leverages LLMs to generate visual symptoms in a scalable manner and employs two learnable prompt modules to facilitate knowledge transfer from CLIP to the medical domain. 3) We conduct extensive experiments on two datasets, and the result demonstrates the strong generalization ability of ViP to medical image analysis.

%% file: Section/method.tex
\input{Figure/fig_pipeline}
\section{Method}
\subsection{Overall Pipeline}
The pipeline of our method is presented in Fig.~\ref{fig:vip}. We consider an input image x and a set of disease labels $C = \{c_1, c_2,..., c_n\}$, where we denote $N$ as the total number of disease categories, with 
$N = n$. The process begins by passing $x$ through a pre-trained vision encoder in the dual-prompt network to compute a feature vector $f$. In parallel, several visual symptoms are generated by the visual symptom generator (VSG) for each disease category. These visual symptoms then undergo transformation in the context prompt module (CoP) to create textual input embeddings for the dual-prompt network. These textual embeddings are then processed through the pre-trained text encoder to compute the textual features for each visual symptom. Next, the merge prompt module (MeP) aggregates text features to obtain a representative feature  $s^c$ for disease category $c$. Going over all categories $c \in C$, we obtain a set of aggregated visual descriptive features $S = \{s^{c_1}, s^{c_2},..., s^{c_n}\}$. Finally, we predict the disease category with the highest cosine similarity score $f \cdot s^c, c \in C$. In the following sections, we will explain the VSG and the dual-prompt network in detail.

\subsection{Visual Symptom Generator (VSG)}
VSG aims to generate a comprehensive set of visual symptoms specific to each disease category. Impressed by the broad knowledge possessed by LLMs and that they can be easily queried with natural language, we propose a two-stage process to construct this set by prompting a large language model, such as GPT-4~\cite{achiam2023gpt}. First, we use a text-only prompt to obtain a coarse set of visual symptoms. We prompt the language model with the following text as the input:
\begin{lstlisting} 
Q:  I am going to use CLIP, a vision-language model to detect {category} in {modality}. What are useful medical visual features for diagnosing {category}? Please list in bullet points and explain in plain words that CLIP understands. Avoid using words such as {category}.
\end{lstlisting}
where \{category\} is substituted for a given category $c \in C$ and \{modality\} is substituted for the imaging modality of the dataset, \eg, dermoscopic images. The prompt is designed to provide sufficient background for GPT-4
and ensure the answers are understandable by CLIP. Next, we refine the coarse set by leveraging the visual-question-answering function of GPT-4. We prompt it with multiple images for each disease category using the following query:
\begin{lstlisting} 
Q:  Please provide visual features regarding color, shape, and texture of this {category} image, which contains 16 sub-images.
\end{lstlisting} 
After receiving the response that encompasses a set of commonly observed visual features across images, the refined set is obtained by intersecting the initial coarse set with the response. Fig.~\ref{fig:description} demonstrates the visual symptoms generated by GPT-4~\cite{achiam2023gpt} using our designed pipeline. As expected, generated visual symptoms typically cover descriptions of color and shape of lesions, presence or absence of certain structures, and other relevant visual features.

\subsection{Dual-Prompt Network}
The dual-prompt network is built upon CLIP. We freeze the image encoder and text encoder of CLIP to retain the general knowledge from the large-scale pre-training data. Unlike conventional CLIP-based approaches that rely on category names for textual input, we use visual symptoms generated from the VSG to enable the model to facilitate the alignment of image features with visual descriptive features. However, the generalization ability of our framework is still limited. This limitation arises due to potential deviations from the expected CLIP text input format in the response from LLMs, and the inherent challenge of effectively aggregating visual symptoms into a disease representation without explicit training~\cite{markovic2007malignant,franquet2001imaging}. Therefore, we further propose two learnable prompt modules: \emph{context prompt} (CoP) and \emph{merge prompt} (MeP), to enhance the model generalization ability. 

\noindent\textbf{CoP.} In addition to category names, context words help to form a complete sentence that specifies the context of the image, which plays a crucial role in the textual input of CLIP. For example, CLIP prepends category names with the context \{a photo of a\}. Similarly, it is desirable to prepend visual symptoms with a customized template to capture the context of medical tasks. However, it is challenging to design hand-crafted templates for visual symptoms due to their more complex phrase structure. Motivated by~\cite{zhou2022learning}, we introduce a set of learnable tokens $\{p_i\}_{i=1}^{M}$, where $p_i \in \mathbb{R}^{d}, i=1,2,...M$, and $d$ is the text embedding dimension, before visual symptoms to automatically learn the context of medical tasks in a data-driven manner. Specifically, given a category $c \in C$, and a visual symptom word embedding $e_d$, the final textual input word embedding $T$ for the text encoder is the concatenation of the learnable tokens and $e_d$, which can be formulated as $T = \text{Concat}(p_1,p_2...p_M, e_d)$. 

\noindent\textbf{MeP.} After processing visual symptoms via text encoder, the next step is to merge visual symptoms into a single representation. Previous methods~\cite{liu2023chatgpt,menon2022visual,byra2023few} adopt the average function, which treats all visual symptoms as equally important, or the max function, which diagnoses based on the most prominent feature. However, these functions suffer from inherent bias because not all visual symptoms contribute equally to a disease. Additionally, it is impossible to accurately diagnose a disease based solely on the most prominent visual symptom in all cases. Therefore, we introduce a learnable token for each disease category to learn the representative feature of the disease. Specifically, given a category $c \in C$, text features matrix $T = [T_1^c, T_2^c,..., T_k^c]^T$, where $T \in \mathbb{R}^{k \times d}$ and $d$ is the text embedding dimension, which is obtained by processing related visual symptoms through the text encoder, and a learnable grouping token $g \in \mathbb{R}^{d}$, we first project $g$ and $T$ into query $Q \in \mathbb{R}^{d}$ and key $K \in \mathbb{R}^{k \times d}$ with different weights $W_q \in \mathbb{R}^{d \times d}$ and $W_k \in \mathbb{R}^{d \times d}$, which can be formulated as:
\begin{equation}
Q = gW_q, K = TW_k.
\end{equation}
The aggregated feature $s^c$ is calculated by combining the grouping prompt $g$ and weighted text features matrix $T$, which can be formulated as:
\begin{equation}
s^c = g + \text{Softmax}(\frac{QK^T}{\sqrt{d}})T.
\end{equation}

After obtaining the aggregated visual descriptive features of all disease categories, CoP and MeP are jointly optimized with a cross-entropy loss, which can be formulated as:
\begin{equation}
L_{ce} = -\log\frac{\exp(f \cdot s^{c_y} 
 / \gamma)}{\sum_{i=1}^{N}\exp(f \cdot s^{c_i}  / \gamma)},
\end{equation}
where $c_y$ denotes the ground truth disease category and $\gamma$ is a learned temperature.

\input{Figure/fig_description}

%% file: Figure/fig_pipeline.tex
\begin{figure}[!t]
\centering
\includegraphics[width=\textwidth]{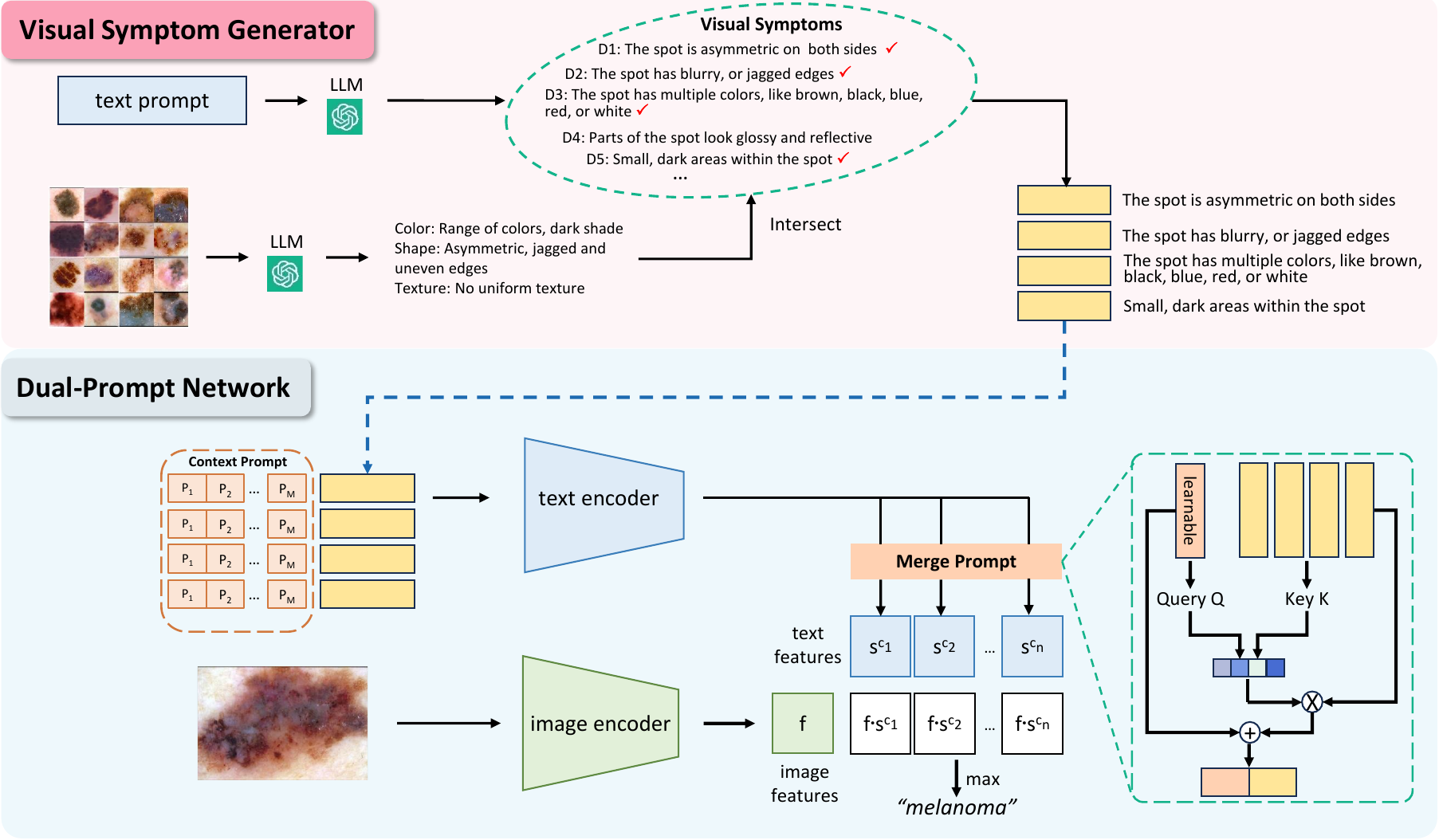}
\caption{Overview of ViP, which consists of a visual symptom generator (VSG) and a dual-prompt network. The visual symptoms predicted by VSG are used as inputs for downstream networks (marked by the blue dashed line).}
\label{fig:vip}
\end{figure}

%% file: Figure/fig_description.tex
\begin{figure}[!t]
\centering
\includegraphics[width=\textwidth]{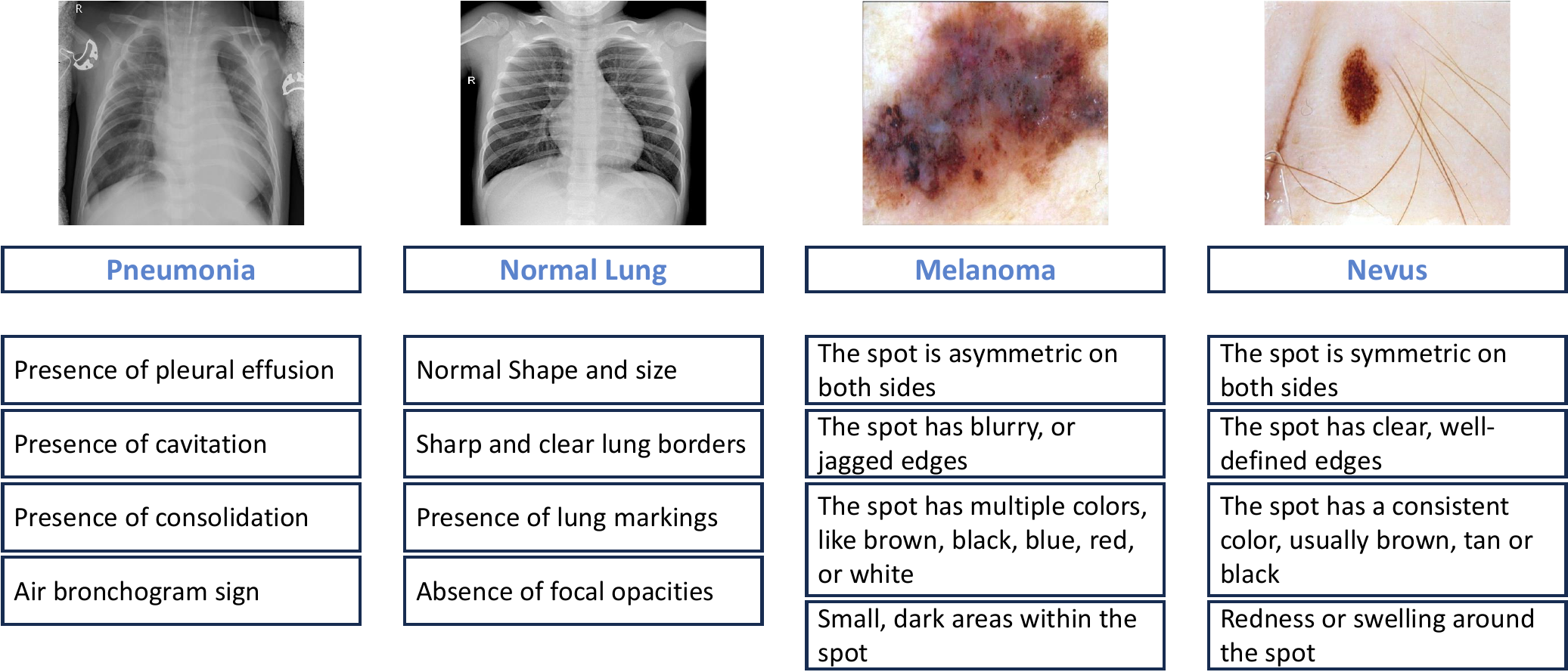}
\caption{Example visual symptoms generated by GPT-4~\cite{achiam2023gpt}.}
\label{fig:description}
\end{figure}

%% file: Section/experiment.tex
\section{Experiments}
\input{Figure/fig_explainability}
\input{Figure/fig_failure}

\subsection{Dataset and Implementation Details}
\textbf{Dataset.} We conduct experiments on two publicly available datasets: Pneumonia~\cite{kermany2018identifying} and Derm7pt~\cite{kawahara2018seven}. Pneumonia consists of chest X-ray images categorized as either normal lung or pneumonia. The official split of this dataset contains 5232 images for training and 624 images for testing. We further randomly divide the training set with a ratio of 9:1 for training and validation. 
Derm7pt consists of over 2000 clinical and dermoscopic images. Following~\cite{patricio2023coherent}, we filter the dataset to obtain 827 images belonging to the  "melanoma" and "nevus" classes, and split the dataset into 346, 161, and 320 images for training, validation, and testing, respectively. For both datasets, we adopt Accuracy (ACC) and Macro F1-score (F1) as evaluation metrics. Macro F1-score addresses the data imbalance issue by computing the arithmetic mean of all per-class F1 scores. 

\noindent\textbf{Implementation Details.} We train the proposed ViP model on an NVIDIA RTX 3090 GPU. Throughout the experiments, we average the results of three vision backbones in CLIP~\cite{radford2021learning}, \ie, ViT-B/16~\cite{dosovitskiy2020image}, ViT-L/14~\cite{dosovitskiy2020image}, and ResNet-50~\cite{he2016deep}. We follow CLIP~\cite{radford2021learning} to set the text embedding dimension $d$ to 512. We follow CoOp~\cite{zhou2022learning} to learn a unified task context and set the length $M$ of the context prompt (CoP) to 4. Training is done with SGD and an initial learning rate of 0.001. The training epoch is set to 50. We follow CLIP~\cite{radford2021learning} to set the temperature $\gamma$ in the cross-entropy loss to $\frac{1}{100}$.

\input{Table/SOTA}

\subsection{Comparisons with State-of-the-art Methods}
\textbf{Effectiveness of explainable visual symptoms.} We conduct a zero-shot experiment to evaluate the effectiveness of visual symptoms for disease diagnosis, while also providing explanations for the decisions. Specifically, our approach makes decision by comparing images to the average embedding of visual descriptive features. As shown in Fig.~\ref{fig:explain}(I), compared with zero-shot CLIP, our method achieves 0.44\% and 18.73\% accuracy improvement, and F1-score gains of 1.58\% and 10.98\% over Pneumonia~\cite{kermany2018identifying} and Derm7pt~\cite{kawahara2018seven}, respectively. This suggests that LLMs can provide useful knowledge for the medical domain. We further analyzed cases where our method correctly predicts the disease category while CLIP fails, as shown in Fig.~\ref{fig:explain}(II).  Our framework improves diagnosis accuracy due to the relatively higher similarity
between the images and the characteristics of the correct category. For instance, Fig.~\ref{fig:explain}(II)(d) is diagnosed as nevus because it demonstrates higher similarity with several characteristics of nevus such as clear edges, consistent brown color, and swelling around the lesion, despite there being a small, black area inside the lesion. 
However, as shown in Fig.~\ref{fig:failure}, there are instances where our method fails to predict the disease category. In Fig.~\ref{fig:failure}(I), although the image exhibits higher similarity with pneumonia characteristics, such as the presence of pleural effusion and air bronchogram sign, the average similarity is lower due to the less obvious symptoms of cavitation and consolidation. This highlights the limitation of using the average function to represent the overall visual features of a disease. In Fig.~
\ref{fig:failure}(II), our method fails to diagnose correctly because the image shares high similarity with nevus characteristics, such as brown color.

\noindent\textbf{Comparison with related methods.} We further compare ViP with several SOTA prompt-based models to evaluate the generalization ability. As shown in Table~\ref{tab:full_data}, ViP achieves highest accuracy of 86.69\%, 81.11\%, and F1-score of 84.94\% , 77.3\% on Pneumonia~\cite{kermany2018identifying} and Derm7pt~\cite{kawahara2018seven}, respectively, indicating the strong generalization ability of our method. Moreover, compared with the fully supervised learning mode, ViP achieves competitive result on Pneumonia~\cite{kermany2018identifying} , but outperforms a great margin on Derm7pt~\cite{kawahara2018seven} where there is less training data, demonstrating the strong generalization ability of ViP in low-resource settings.

\input{Table/abl}
\input{Figure/fig_knowledge}
\subsection{Ablation Study}
\textbf{Effectiveness of each component.} We conduct ablation studies to explore the effectiveness of each component in ViP, as shown in Table~\ref{tab:abl}. Compared with zero-shot baseline, both the integration of CoP and MeP exhibit considerable improvement, demonstrating the importance of learning medical task context and effective aggregation of visual symptoms. Moreover, compared with non-parametric aggregation methods, such as average and max functions~\cite{byra2023few,menon2022visual}, our proposed MeP outperforms in both datasets. This result further validates the effectiveness of our method.
\\
\textbf{Knowledge Faithfulness.} We conduct an additional experiment to validate our argument that LLM-generated visual symptoms provide useful knowledge for the generalization to the medical domain. As shown in Fig.~\ref{fig:knowledge}, we replace the visual symptoms of nevus with three types of knowledge: 1) Out-of-domain knowledge, involving visual symptoms unrelated to the medical domain, such as descriptions of food. 2) Useless knowledge, referring to descriptions associated with our target disease but do not provide useful information for diagnosis, such as descriptions of skin structure. 3) Incorrect knowledge, which provides erroneous symptoms for diagnosis. In this experiment, we alter certain words in the descriptions to their antonyms to create misleading descriptions of nevus. Compared to other variations, LLM-generated knowledge achieves best performance, indicating that accurate visual symptoms contribute to the generalization in the medical domain. 

%% file: Figure/fig_explainability.tex
\begin{figure}[!t]
\centering
\includegraphics[width=\textwidth]{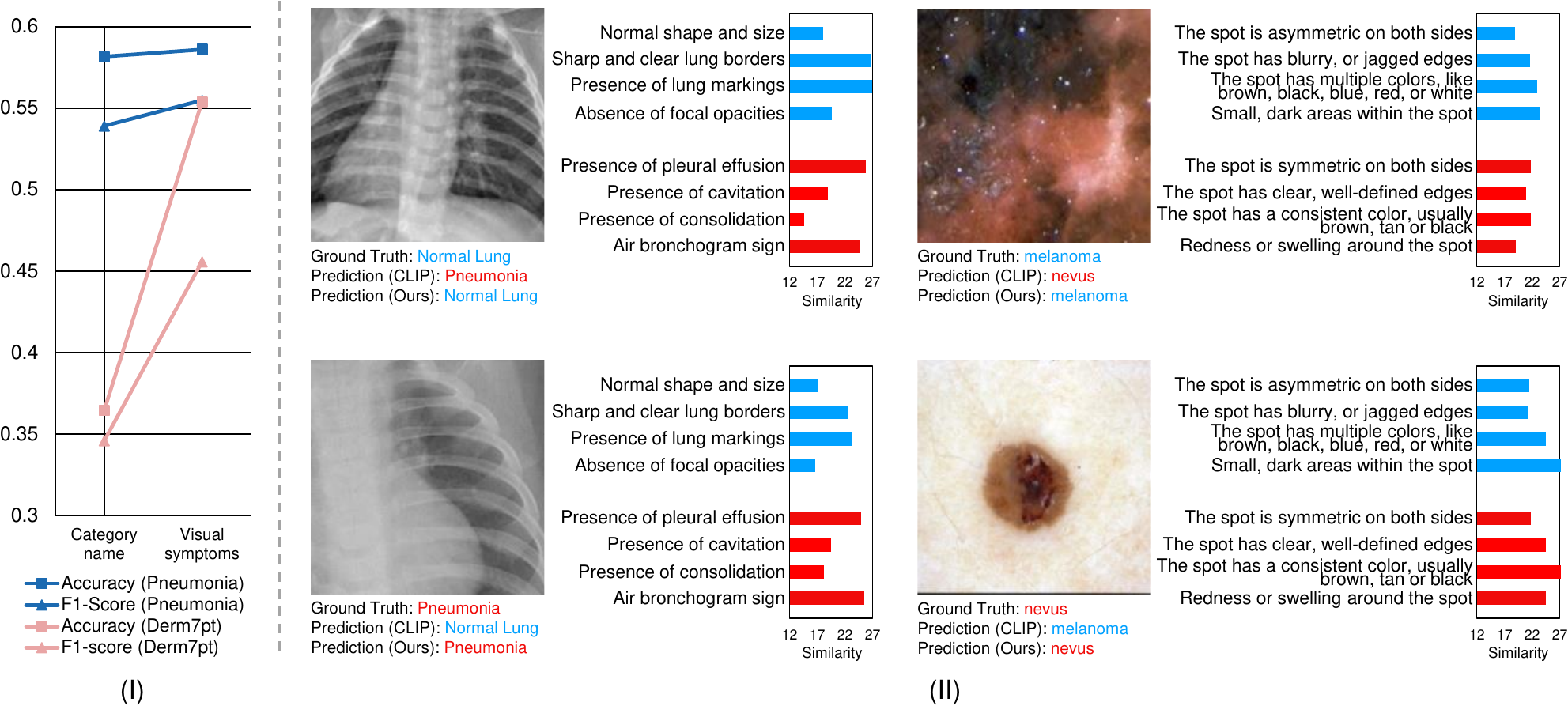}
\caption{(I) Zero-shot CLIP with category name or visual symptoms as text inputs. (II) Diagnostic process based on cosine similarity scores between images and visual symptoms.
}
\label{fig:explain}
\end{figure}

%% file: Figure/fig_failure.tex
\begin{figure}[!t]
\centering
\includegraphics[width=\textwidth]{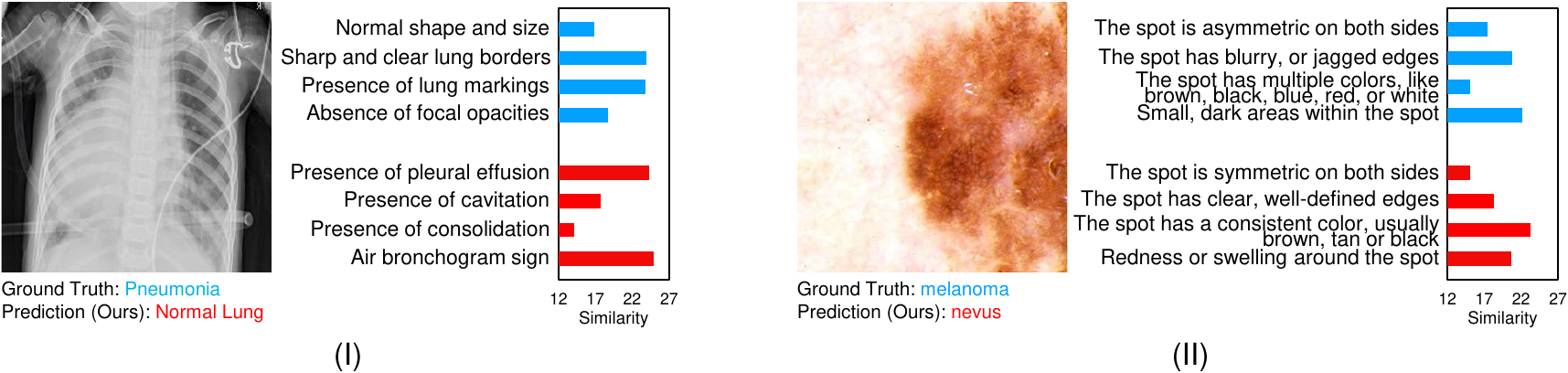}
% \caption{Cases of misclassification in zero-shot experiment.}
\caption{Failure cases in the zero-shot experiment.}
\label{fig:failure}
\end{figure}

%% file: Table/SOTA.tex
\begin{table}[!t]
\centering
\caption{Result comparisons with SOTAs. The mean and standard deviation is computed across three vision backbones.}
\setlength{\tabcolsep}{8pt}{
    \begin{tabular}{c cc cc}
    % \hline
    \toprule
    \multirow{2}{*}{Method} & \multicolumn{2}{c}{Pneumonia}      & \multicolumn{2}{c}{Derm7pt} \\ 
    & {ACC} & {F1} & {ACC} & {F1} \\
    % \hline
    \midrule
    CoOp~\cite{zhou2022learning} & 0.8337$_{0.019}$ & 0.8148$_{0.017}$ & 0.7823$_{0.005}$ & 0.7328$_{0.017}$ \\
    CoCoOp~\cite{zhou2022conditional} & 0.8440$_{0.025}$ & 0.8217$_{0.032}$ & 0.7668$_{0.014}$ & 0.6647$_{0.057}$ \\
    KgCoOp~\cite{yao2023visual} & 0.8303$_{0.022}$ & 0.8010$_{0.027}$ & 0.7726$_{0.009}$ & 0.7093$_{0.033}$ \\
    Bayesian~\cite{derakhshani2023bayesian} & 0.8301$_{0.041}$ & 0.8081$_{0.048}$ & 0.6921$_{0.014}$ & 0.5561$_{0.054}$ \\
    MaPLe~\cite{khattak2023maple} & 0.8553$_{0.034}$ & 0.8393$_{0.036}$ & 0.7903$_{0.038}$  & 0.7250$_{0.073}$  \\
    \midrule
    Supervised & 0.8660$_{0.025}$ & 0.8530$_{0.025}$ & 0.7277$_{0.044}$ & 0.6236$_{0.093}$ \\
    \midrule
    ViP$_{\textrm{ours}}$ & \textbf{0.8669}$_{0.031}$ & \textbf{0.8494}$_{0.036}$ & \textbf{0.8111}$_{0.007}$
    & \textbf{0.7730}$_{0.015}$  \\
    \bottomrule
    \end{tabular}}
\label{tab:full_data}
\end{table}

%% file: Table/abl.tex
\begin{table}[!t]
\centering
\caption{Ablation study results. ``Context'' and ``Merge'' denote context prompt (CoP) and merge prompt (MeP), respectively. ``Max'' and ``Mean'' denote the maximum and average of visual descriptive features, respectively.}    
\setlength{\tabcolsep}{15pt}{
\begin{tabular}{cc  cc cc}
 \toprule
    \multirow{2}{*}{Context} & \multirow{2}{*}{Merge} & \multicolumn{2}{c}{Pneumonia}  & \multicolumn{2}{c}{Derm7pt} \\ 
    & & 
    {ACC} & {F1} & {ACC} & {F1} \\
    \midrule
    \xmark & \xmark & 0.5861 & 0.5549 & 0.5539 & 0.4558 \\
    \xmark &  \cmark & 0.8486 & 0.8312 & 0.7531 & 0.6194 \\
    \cmark & Max & 0.8390 & 0.8223 & 0.7970 & 0.7506 \\
    \cmark & Mean & 0.8550 & 0.8347 & 0.8041 & 0.7646 \\
    \cmark & \cmark & \textbf{0.8669} & \textbf{0.8494} & \textbf{0.8111} & \textbf{0.7730} \\
    \bottomrule
\label{tab:abl}  
\vspace{-3mm}
\end{tabular}}
\end{table}

%% file: Figure/fig_knowledge.tex
\begin{figure}[!t]
\centering
\includegraphics[width=\textwidth]{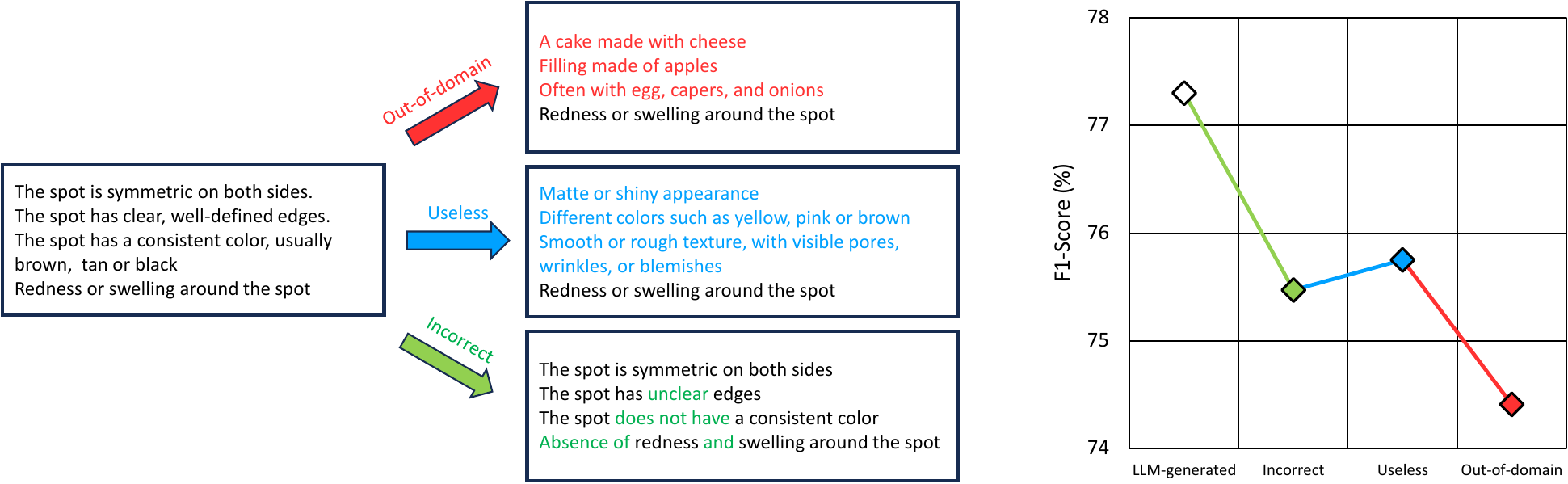}
\caption{Ablation study comparing with different types of knowledge.}
\label{fig:knowledge}
\end{figure}

%% file: Section/conclusion.tex
%\section{Conclusion, Limitations and Future Work}
\section{Conclusion}
This paper presented a novel visual symptom-guided prompt learning pipeline, referred to as ViP, which effectively transfers knowledge from VLMs to medical image analysis. By leveraging pre-trained LLMs, ViP generates useful visual symptoms to guide CLIP in aligning image features with visual symptoms. Additionally, ViP incorporates two learnable prompt modules, context prompt and merge prompt, to further enhance the generalization ability. Experimental results underscored the effectiveness of each module and the superior performance of our pipeline to state-of-the-art methods. Future work will focus on extending the framework to other medical image analysis tasks, such as the diagnosis of rare diseases and malformed organs, where data and annotations are scarce and costly. Additionally, we will investigate techniques to enhance the interpretability of context prompt.

\begin{credits}
\subsubsection{\ackname} We thank Yi Gu, Yibo Hu and Xiaoyu Fu for their helpful discussions. This work was supported by the Hong Kong Innovation and Technology Fund (Project No. MHP/002/22), Project of Hetao Shenzhen-Hong Kong Science and Technology Innovation Cooperation Zone (HZQB-KCZYB-2020083) and the Research Grants Council of the Hong Kong (Project Reference Number: T45-401/22-N).

\subsubsection{\discintname}
The authors have no competing interests to declare that are
relevant to the content of this paper.
\end{credits}